\def\year{2022}\relax
\documentclass[letterpaper]{article} 
\usepackage{aaai24}  
\usepackage{times}  
\usepackage{helvet}  
\usepackage{courier}  
\usepackage[hyphens]{url}  
\usepackage{graphicx} 
\def\UrlFont{\rm}  
\usepackage{natbib}  
\usepackage{caption} 
\DeclareCaptionStyle{ruled}{labelfont=normalfont,labelsep=colon,strut=off} 
\usepackage{algorithm}
\usepackage{algorithmic}
\usepackage{amsmath}
\usepackage{amsfonts}
\usepackage{dblfloatfix}
\usepackage{float}
\usepackage{newfloat}
\usepackage{listings}
\usepackage{subfig}
\floatstyle{ruled}
\newfloat{listing}{tb}{lst}{}
\floatname{listing}{Listing}

\usepackage{xcolor}

\nocopyright
\title{PULSAR: Graph based \underline{P}ositive \underline{U}nlabeled \underline{L}earning with Multi \underline{S}tream \underline{A}daptive Convolutions for Parkinson's Disease \underline{R}ecognition}
\author{
    Md. Zarif Ul Alam \textsuperscript{\rm 1},
    Md Saiful Islam \textsuperscript{\rm 2},
    Ehsan Hoque \textsuperscript{\rm 2},
    M Saifur Rahman \textsuperscript{\rm 1}
}
\affiliations{
    \textsuperscript{\rm 1} Bangladesh University of Engineering and Technology, Bangladesh\\
    \textsuperscript{\rm 2} University of Rochester, United States\\

}

\usepackage{bibentry}

\begin{document}

\maketitle

\setlength{\textfloatsep}{5pt plus 1.0pt minus 2.0pt}

\begin{figure*}[h]
  \begin{center}
  \includegraphics[width=\linewidth]{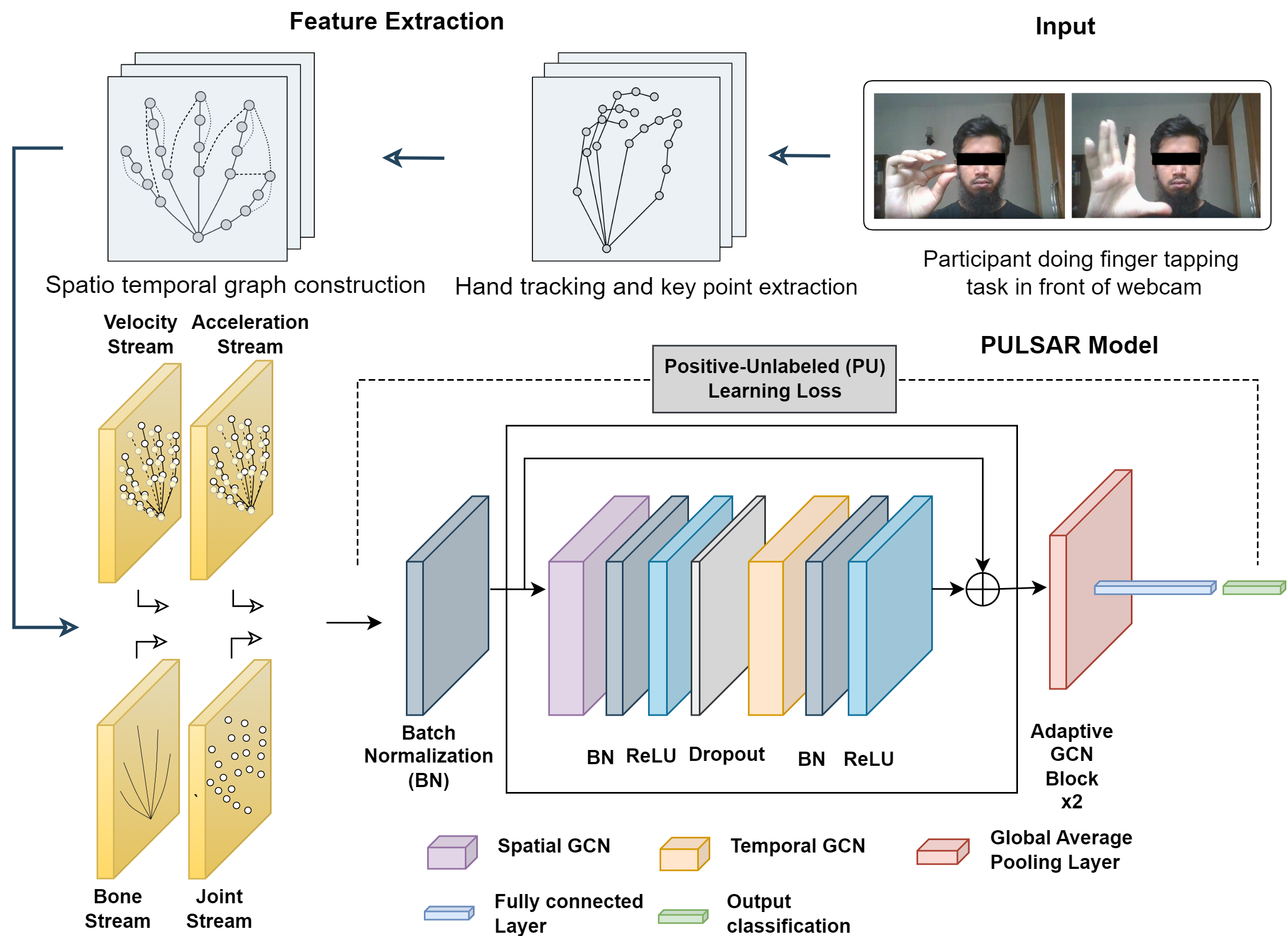}
  \caption{Overview of the PD screening pipeline. A participant can perform finger-tapping task in front of a computer webcam. A hand tracking model is used to locate the key points of the hand. A spatio temporal graph is constructed specifically for the finger-tapping task. Four different feature streams (joint, bone, velocity and acceleration) are generated and fed to the proposed PULSAR model for prediction.}
  \label{fig:stgcn_full_overview}
  \end{center}
  \vspace{-4mm}
\end{figure*}
\begin{abstract}
Parkinson's disease (PD) is a neuro-degenerative disorder that affects movement, speech, and coordination. Timely diagnosis and treatment can improve the quality of life for PD patients. However, access to clinical diagnosis is limited in low and middle income countries (LMICs). Therefore, development of automated screening tools for PD can have a huge social impact, particularly in the public health sector. In this paper, we present PULSAR, a novel method to screen for PD from webcam-recorded videos of the finger-tapping task from the Movement Disorder Society - Unified Parkinson’s Disease Rating Scale (MDS-UPDRS). PULSAR is trained and evaluated on data collected from 382 participants (183 self-reported as PD patients). We used an adaptive graph convolutional neural network to dynamically learn the spatio temporal graph edges specific to the finger-tapping task. We enhanced this idea with a multi stream adaptive convolution model to learn features from different modalities of data critical to detect PD, such as relative location of the finger joints, velocity and acceleration of tapping. As the labels of the videos are self-reported, there could be cases of undiagnosed PD in the non-PD labeled samples. We leveraged the idea of Positive Unlabeled (PU) Learning that does not need labeled negative data. Our experiments show clear benefit of modeling the problem in this way. PULSAR achieved 80.95\% accuracy in validation set and a mean accuracy of 71.29\% (2.49\% standard deviation) in independent test, despite being trained with limited amount of data. This is specially promising as labeled data is scarce in health care sector. We hope PULSAR will make PD screening more accessible to everyone. The proposed techniques could be extended for assessment of other movement disorders, such as ataxia, and Huntington’s disease.
\end{abstract}

\vspace{-4.5mm}

\section{Introduction}


Parkinson’s disease (PD) is one of the common degenerative diseases of the nervous system \cite{ji2018filtering}. It is characterized by a variety of life-changing motor dysfunction symptoms, including tremor, Bradykinesia (slowness of movement), rigidity (limb stiffness), impaired balance and gait, etc. Over 10 million people worldwide are affected by PD \cite{intro_wiki}. In 2016, PD resulted in about 211,000 deaths globally, an increase of 161\% since 1990. 
The diagnosis of PD mainly relies on clinical criteria based on the Parkinsonian symptoms (e.g., tremor, Bradykinesia), and medical history. However, the clinical diagnostic is challenged by the subjective opinions or experiences of different medical experts. In addition, it is not accessible to many individuals since the number of neurologists is very limited in some countries \cite{kissani2022does}. Therefore, an efficient and remote automatic PD diagnosis system is valuable for supporting clinicians with more robust diagnostic decision-making. 

With the advancement in computer vision technology, sensing of subtle movement of face and body has become possible, and this has the potential to lead to important medical and physiological implications. However, healthcare data (specially, data from individuals with PD) is hard to obtain and the small size of the datasets limits application of state-of-the art machine learning models. Here we introduce PULSAR -- a graph neural network based deep learning architecture that can be trained with limited data. In addition, in a dataset that uses patient-reported PD diagnosis as the ground truth, ``negative'' data samples remain unreliable -- many people live with PD without being diagnosed. 
In context of healthcare this is very common. Rather than expecting clean data, we focused on dealing with the problem objectively. To the best of our knowledge, we are the first to explore Positive Unlabeled (PU) learning in the context of PD screening, which eliminates the need for reliable ``negative'' samples. An overview of the PD screening pipeline leveraging PULSAR model is shown in Figure~\ref{fig:stgcn_full_overview}. PULSAR has been tested through a comprehensive set of experiments, the results of which support its efficacy. As many movement disorders (e.g., ataxia, Huntington's disease) share similarity in terms of the tasks used for screening/diagnosis and data collection, our proposed approach could be applied for those diseases, potentially improving access to neurological care. In this regard, the data processing and model training codes and the processed de-identified dataset will be made publicly available upon acceptance of the paper. 

In summary, we make the following contributions:

\begin{itemize}
    \item We propose a novel multi stream adaptive convolution model which allows PD screening from videos of the finger-tapping task with limited data.
    
    \item We propose PU learning for PD screening that eliminates the necessity of labeled negative data for model training.
    
    \item The model achieved 80.95\% and 71.29\% accuracy on the validation and test set, respectively, demonstrating the efficacy of our proposed approach.
\end{itemize}

\vspace{-5mm}
\section{Related Work}
Clinical diagnosis of PD involves a comprehensive assessment of a patient's medical history, neurological examination, and the presence of characteristic motor features. Numerous clinical studies focus on identifying motor symptoms, analyzing brain imaging data, and investigating physiological biomarkers to diagnose PD. 
For example, neuroimaging techniques have furthered the understanding of PD pathophysiology. PET (Positron Emission Tomography) and SPECT (Single Photon Emission Computed Tomography) scans have been used to visualize dopamine transporter binding in the basal ganglia, aiding in differential diagnosis and disease progression assessment \cite{marek1996sup}. These clinical approaches provide essential groundwork for understanding PD and serve as a reference for developing computational methods. But the clinical diagnosis method is not cost-effective and as such it is very hard to make PD diagnosis accessible for all.

Machine Learning (ML) algorithms have recently been used for PD detection and diagnosis tasks~\cite{senturk2020early, li2022detecting, govindu2023early} due to their powerful predictive performance. Studies have shown that ML can provide early and reliable diagnosis of patients and help doctors make decisions by analyzing speech, gait or motor tasks. 
In parallel to traditional ML techniques, Deep Learning (DL) techniques have attracted a lot of attention because of their powerful automatic feature extraction capabilities \cite{shahid2020deep,srinidhi2021deep}. For example, \citeauthor{sivaranjini2020deep} developed a deep learning-based approach that classifies PD using Magnetic Resonance (MR) images, and \citeauthor{johri2019parkinson} utilized neural network-based techniques to detect PD from gait and speech, showcasing the potential of DL methods. A common problem with many of these models is that they do not make diagnosis easily available to everyone. For example, Neuroimaging or MRI based models, despite performing well, are expensive and also intrusive. Speech-based models have limitations too, as they are not easily generalizable because the language and pronunciation habits of people in different countries vary significantly. 
Also, in most cases, the training set is too small that these models do not generalize well.

Our work is inspired by recent developments in Graph Neural Networks (GNNs). GNN has emerged as a powerful tool for analyzing complex and structured data, making it particularly valuable in the context of healthcare and medical data analysis. \citeauthor{yan2018spatial} and \citeauthor{shi2019two} achieved remarkable performance in skeleton based action recognition using graph convolutional networks. \citeauthor{kim2021interpretable} introduced a GNN-based approach for predicting disease progression in Alzheimer's patients using brain connectivity graphs based on longitudinal neuroimaging data. \citeauthor{zhao2022attention} used an attention-based graph neural network (AGNN) for early diagnosis of PD via diffusion tensor imaging (DTI) data and phenotypic information. In addition, PU learning enables a classifier to learn from positive and unlabeled samples \cite{qiu2009building, nguyen2011positive}, which is particularly significant for medical data analysis, as the absence of a disease is often costly to verify. 
\vspace{-3.5mm}
\section{Dataset}
The dataset contains recorded videos from $382$ unique participants completing the finger-tapping task in front of a computer webcam. The finger-tapping task involves repeatedly tapping the thumb-finger with the index-finger (10 times) as fast and as big as possible. The participants completed the task twice, once with the right hand, and another with the left hand. Each participant was asked to self-report whether they were clinically diagnosed with Parkinson's disease or not. Data was collected using the PARK tool~\cite{langevin2019park}, a public website\footnote{\url{https://parktest.net}} accessible from computers with major Internet browsers.

The training and validation dataset consists of recorded vidoes from $200$ unique participants. Out of them $100$ are self reported PD patients. 20\% of this set is used for validation purpose. The independent test set has videos from $182$ unique participants different from the training and validation set. Out of them $83$ are self reported PD patients. Each of the videos were segmented into clips of 80 frames and each of this clips are treated as one sample of our dataset. Videos from individuals with self-reported PD were labeled as positive samples, while individuals not reporting a diagnosed PD remained unlabeled. Table \ref{tab:demography} presents demographic information of the study participants. It is notable that, due to low prevalence of PD (in 2022, the overall prevalence of PD among persons older than 45 was 0.572 per 100~\cite{willis2022incidence}), typically, the unlabeled cohort are considered to not have PD~\cite{islam2023using}. However, it is possible that some of the unlabeled participants did have PD, but remained undiagnosed. This makes positive unlabeled learning particularly suitable for this problem.

\begin{table}[t]
\centering
\resizebox{0.90\columnwidth}{!}{%
\begin{tabular}{|lllll|}
\hline
\multicolumn{2}{|l}{\textbf{Characteristics}}                                                 & \textbf{With PD}      & \textbf{Unlabeled}  & \textbf{Total}                 \\ \hline
\multicolumn{2}{|l}{\begin{tabular}[c]{@{}l@{}}No. of \\ Participants, n \end{tabular} }                                  & 183 & 199 & \textbf{382}  \\ \hline
\multicolumn{5}{|l|}{Sex, n (\%)}                                                                                                      \\
 & Male                                                                              & 112 (61.2\%) & 77 (38.7\%) & \textbf{189 (49.5\%)} \\
 & Female                                                                            & 69 (37.7\%)  & 122 (61.3\%) & \textbf{191 (50.0\%)} \\ 
 & Not available                                                                            & 2 (1.1\%)  & 0 (0.0\%) & \textbf{2 (0.5\%)} \\ \hline
\multicolumn{5}{|l|}{Age in years, n (\%)}                                                                                 \\
 & Below 20                                                                     & 1 (0.5\%)    & 4 (2.0\%)   & \textbf{5 (1.3\%)}    \\
 & 20-39                                                                             & 4 (2.2\%)    & 20 (10.0\%)  & \textbf{24 (6.3\%)}   \\
 & 40-59                                                                             & 48 (26.2\%)    & 47 (23.6\%)   & \textbf{95 (24.9\%)}    \\
 & 60-79                                                                             & 119 (65.0\%)    & 126 (63.3\%)   & \textbf{245 (64.1\%)}   \\
 & Above 80                                                                             & 9 (4.9\%)  & 1 (0.5\%)  & \textbf{10 (2.6\%)}  \\
 & Not available                                                                             & 2 (1.1\%)  & 1 (0.5\%)  & \textbf{3 (0.8\%)}  \\\hline
\end{tabular}%
}
\caption{Demographic characteristics of the participants. With PD column represents the participants who self-reported to be diagnosed with PD, while the rest of the participants remained unlabeled.
}
\label{tab:demography}
\end{table}

\color{black}
\vspace{-1.5mm}
\section{Methodology}
Our developed method PULSAR focuses on the finger-tapping task. This is commonly used in neurological exams to evaluate bradykinesia (i.e., slowing of movement) in upper extremities, which is a key symptom of PD \cite{hughes1992accuracy}. The input consists of video recordings of each participant sitting in front of a webcam and doing the finger-tapping task.

\subsubsection{Video Augmentation}
In our experiments, we employed three types of video augmentation to increase the size and diversity of our dataset. These are horizontal flip, vertical flip, and horizontal-vertical flip of the input videos. These transformations simulate variations in camera position and hand orientation and were applied to each video in the dataset. By flipping the videos, the size of the dataset became fourfold the original size. We also normalize the pixel intensities of each frame and resized them to a fixed resolution to ensure that the model is trained on consistent inputs.

\subsubsection{Data Cleaning}
\begin{figure*}[htb]
  \centering
  \includegraphics[width=\linewidth]{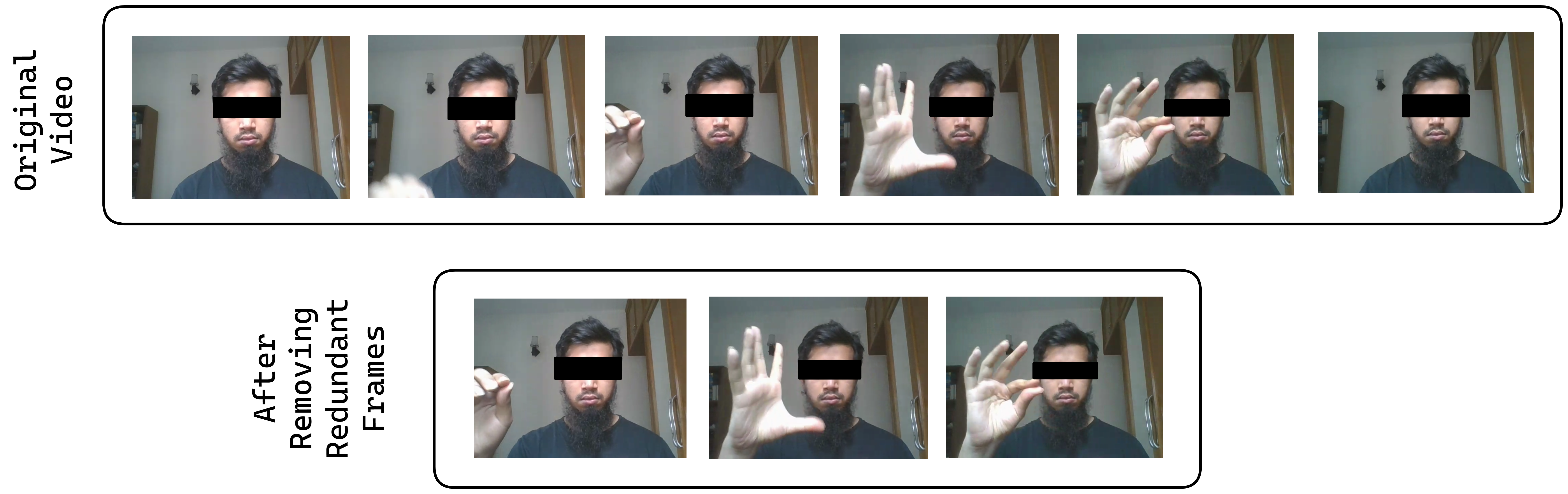}
  \caption{Demonstration of the described data cleaning process.\footnotemark}
  \label{fig:data_clean}
  \vspace{-3mm}
\end{figure*}

Data cleaning was performed to prevent learning from irrelevant or noisy data. We removed frames where no hand was visible -- typically at the beginning and end of the videos when the subject was preparing to start the task or lowering the hand after completing the task (Figure ~\ref{fig:data_clean}). To this end, we used MediaPipe Hands~\cite{zhang2020mediapipe} to detect the presence of a hand in each frame, and removed the frames where no hand was visible. We were thus left with a clean and consistent dataset containing only relevant frames. This step proved crucial as the model subsequently became robust, with a considerable boost in performance (11\% increase in accuracy in validation set). Furthermore, this expedited our training process because the number of frames got considerably reduced.

\subsubsection{Hand Key Point Extraction}
We extracted hand key point coordinates from RGB video data using MediaPipe that offers high accuracy and fast inference speed. Twenty one 3-D hand key points were extracted, including four joints per finger and one wrist joint, following the structure shown in Figure~\ref{fig:all_in_one}. The minimum detection confidence and minimum tracking confidence were empirically set to 0.8 and 0.9 respectively to reduce false positives.

\subsubsection{Spatio Temporal Graph Construction}
\begin{figure}[htb]
  \centering
  \subfloat[]{
    \includegraphics[width=0.47\linewidth]{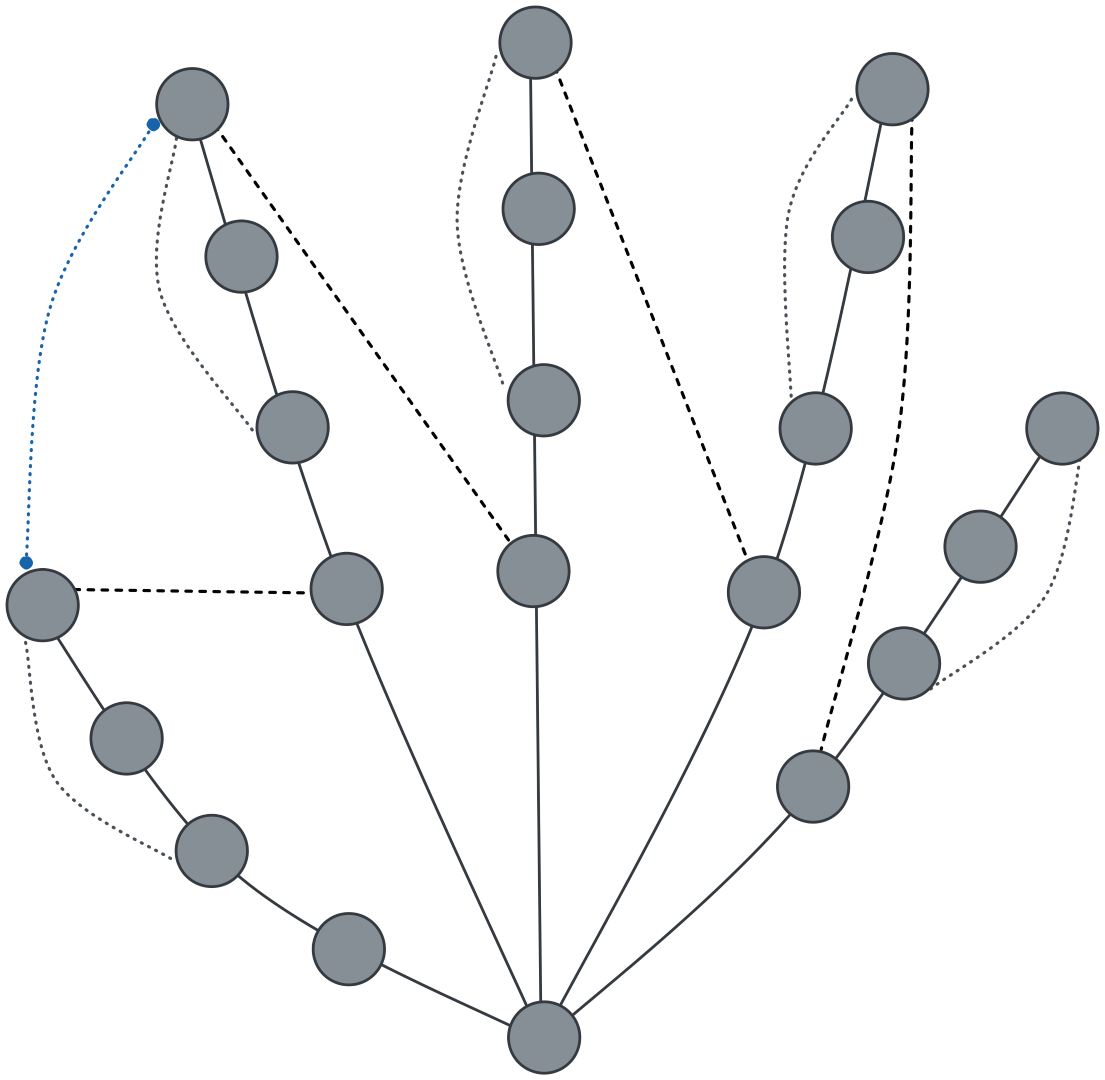}
    \label{fig:hgwe}
  }
  \subfloat[]{
    \includegraphics[width=0.52\linewidth]{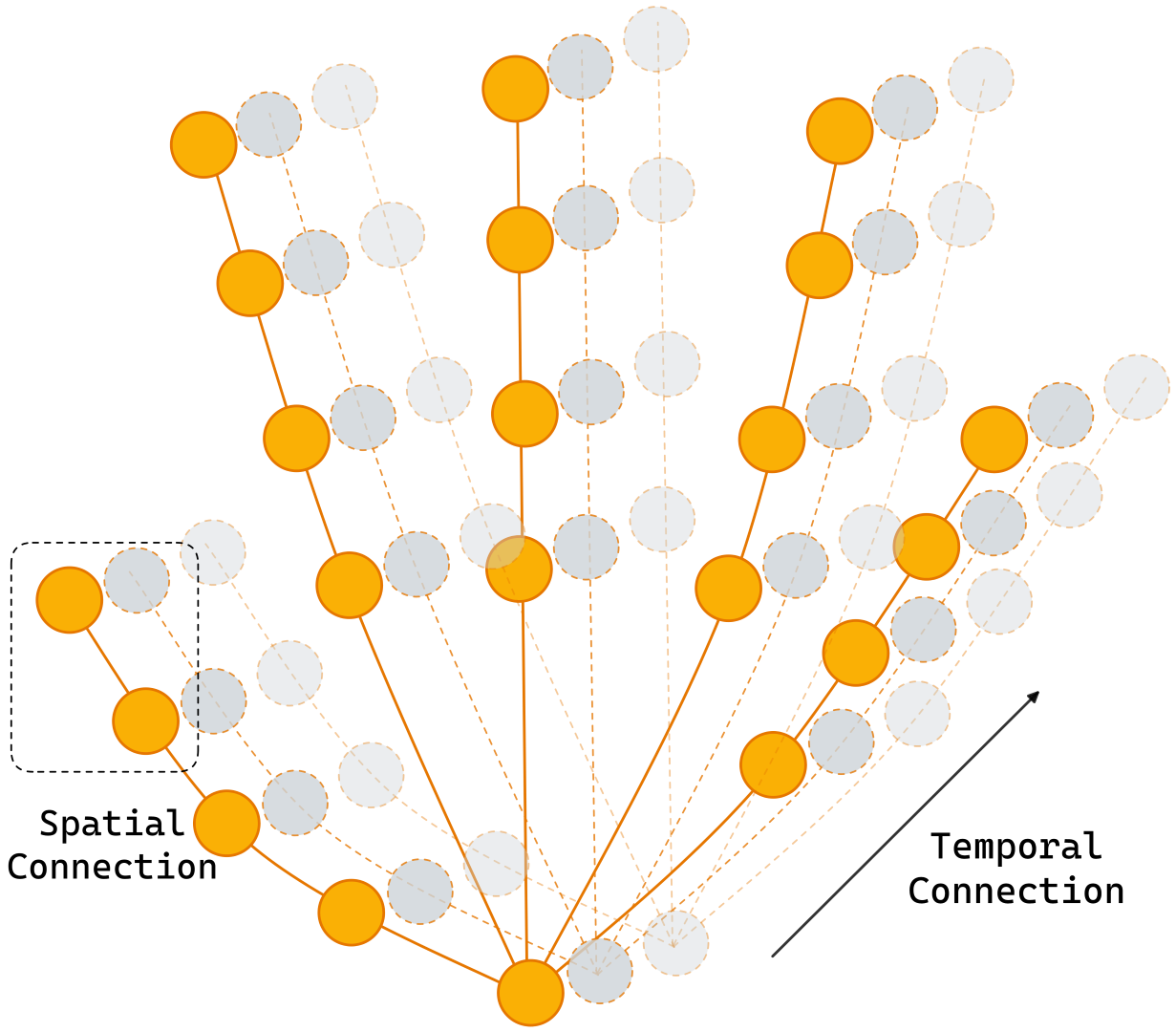}
    \label{fig:spatio}
  }
 \caption[Spatio Temporal Graph Construction]{Spatio Temporal Graph Construction. (a) The spatial connections of the finger and wrist joints. Natural joint connections are denoted by solid lines. The first type of augmented edge is denoted by dashed lines, the second type by dotted lines, and the third by a blue dashed arrowhead. (b) The temporal connections are between the same joints in the successive frames.}
 \label{fig:all_in_one}
 \vspace{-3mm}
\end{figure}

The inherent graph structure connects each joint with its immediate natural neighbor, culminating in a graph with 21 vertices and 20 edges. However, this proposed graph might lack the capacity to capture fine-grained hand movements~\cite{li2019spatial}. For the purpose of obtaining more contextually significant semantic information, we introduced three types of augmented edges (Figure \ref{fig:hgwe}). The first type of edge connects the fingertips to the base of the adjacent finger to the right for the right hand (to the left for the left hand). The second type of edge connects the fingertip to the middle joint of the same finger. The third type establishes a connection between the tip of the thumb and the index finger. These augmented edges are able to capture additional insights into various hand states, such as the horizontal and vertical distance between two fingers and finger bending. The third type of connection is particularly relevant for the finger-tapping task.

\footnotetext{Images are for demonstration purpose and not part of the dataset.}

\subsubsection{Adaptive Graph Convolution}

In the context of the spatio temporal graph convolution network, the graph convolution process can be defined as follows~\cite{yan2018spatial}:
\begin{equation}
\mathbf{f}_{out }=\sum_k^{K_v} \mathbf{W}_k\left(\mathbf{f}_{in} \mathbf{A}_k\right) \odot \mathbf{M}_k
\label{eq:graph-conv-full}
\vspace{-1mm}
\end{equation}

where $\mathbf{f}_{in}$ stands for the input feature map, $\mathbf{f}_{out}$ signifies the resulting feature map, $K_v$ represents the spatial dimension's kernel size, $\mathbf{W}_k$ is the weight vector of the $1 \times 1$ convolution. $\mathbf{M}_k$ is an $N \times N$ attention map that indicates the significance of each vertex. The symbol $\odot$ denotes the dot product operation. The matrix $A_k$ determines the presence of connections between two vertices. For the temporal dimension, where each vertex has a fixed number of neighbors (corresponding to joints in consecutive frames), the execution of graph convolution is akin to the traditional convolution operation, following the approach described by \citeauthor{yan2018spatial}.

In a graph convolutional network, the convolution process involves aggregating data from neighboring nodes within the graph. This combined data, along with the node's inherent characteristics, is subsequently passed through a non-linear activation function to generate the convolutional output. But having a fixed predefined graph might not be optimal for capturing nuanced distinctions in motor tasks performed by Parkinson's patients. Such an approach could limit the capacity of the graph convolutional network (GCN) to acquire comprehensive insights from alternative connections. The fixation of the graph structure throughout all GCN layers contradicts the principle that the neural network could acquire diverse features at each layer. Adaptive graph convolution as presented in \citeauthor{shi2019two}, is a type of graph convolution that allows for learnable weights on the aggregation step of the convolution operation. Unlike traditional GCNs, where the aggregation weights used to combine data from neighboring nodes remain fixed and consistent across all nodes, adaptive graph convolution allows these weights to be learned during the training process. 

The adaptive graph convolution layer optimizes the graph's structure alongside other network parameters through an end-to-end learning approach, as demonstrated by \citeauthor{shi2019two}. Notably, the graph's configuration is distinct for various layers and instances, significantly enhancing the model's adaptability. It is also designed as a residual component, ensuring the stability of the underlying model.

To make the graph structure adaptive, Eq. \ref{eq:graph-conv-full} is changed as: $\mathbf{f}_{\text {out }}=\sum_k^{K_v} \mathbf{W}_k * \mathbf{f}_{i n} \left( \mathbf{A}_k + \mathbf{B}_k + \mathbf{C}_k\right)$ \cite{shi2019two}.
The adaptive spatial graph convolution layer integrates both the provided adjacency matrix ($A_k$) and parameterized, optimized adjacency matrices ($B_k$ and $C_k$). These matrices are detailed as follows.

\begin{itemize}
\item $\mathbf{A_k}$ denotes the original normalized adjacency matrix, capturing inherent physical structure of human hands.
\item $\mathbf{B_k}$ represents a fully learnable and parameterized adjacency matrix. The absence of constraints on the values of $B_k$ permits the graph to be learned based on training data. $B_k$ is capable of generating novel connections, similar to attention mechanism, yet offering greater flexibility due to its ability to establish connections absent in the original physical graph.
\item $\mathbf{C_k}$ signifies a data-dependent adjacency matrix, derived by embedding input features via a $1 \times 1$ convolutional operation and a \texttt{softmax} function. To ascertain the presence and strength of connections between vertex pairs, a normalized embedded Gaussian function computes the similarity between the vertices \cite{shi2019two}.
\end{itemize}

Instead of directly substituting the original $A_k$ with $B_k$ or $C_k$, they are incorporated alongside it. This strategy amplifies model flexibility while preserving performance.

\vspace{-1.5mm}
\subsubsection{Positive Unlabelled Learning}

Let random variables $\boldsymbol{x} \in \mathbb{R}^d$ and $y \in\{+1,-1\}$ be equipped with probability density $p(\boldsymbol{x}, y)$, where $d$ is a positive integer. In a binary classification problem from $\boldsymbol{x}$ to $y$, let's assume there are three sets of samples: positive $(\mathrm{P})$, negative $(\mathrm{N})$, and unlabeled $(\mathrm{U})$.

$$
\begin{aligned}
& \mathcal{X}_{\mathrm{P}}:=\left\{\boldsymbol{x}_i^{\mathrm{P}}\right\}_{i=1}^{n_{\mathrm{P}}} \stackrel{\text { i.i.d. }}{\sim} p_{\mathrm{P}}(\boldsymbol{x}):=p(\boldsymbol{x} \mid y=+1), \\
& \mathcal{X}_{\mathrm{N}}:=\left\{\boldsymbol{x}_i^{\mathrm{N}}\right\}_{i=1}^{n_{\mathrm{N}}} \stackrel{\text { i.i.d. }}{\sim} p_{\mathrm{N}}(\boldsymbol{x}):=p(\boldsymbol{x} \mid y=-1), \\
& \mathcal{X}_{\mathrm{U}}:=\left\{\boldsymbol{x}_i^{\mathrm{U}}\right\}_{i=1}^{n_{\mathrm{U}}} \stackrel{\text { i.i.d. }}{\sim} p(\boldsymbol{x}):=\theta_{\mathrm{P}} p_{\mathrm{P}}(\boldsymbol{x})+\theta_{\mathrm{N}} p_{\mathrm{N}}(\boldsymbol{x}),
\end{aligned}
$$
where $\theta_{\mathrm{P}}:=p(y=+1), \quad \theta_{\mathrm{N}}:=p(y=-1)$ are the class-prior probabilities for the positive and negative classes such that $\theta_{\mathrm{P}}+\theta_{\mathrm{N}}=1$.

Let $g: \mathbb{R}^d \rightarrow \mathbb{R}$ be an arbitrary real-valued decision function for binary classification, and classification is performed based on its sign. Let $\ell: \mathbb{R} \rightarrow \mathbb{R}$ be a loss function such that $\ell(m)$ generally takes a small value for large margin $m=y g(\boldsymbol{x})$. Let $R_{\mathrm{P}}(g), R_{\mathrm{N}}(g), R_{\mathrm{U}, \mathrm{P}}(g)$, and $R_{\mathrm{U}, \mathrm{N}}(g)$ be the risks of classifier $g$ under loss $\ell$ :
$$
\begin{array}{rrr}
R_{\mathrm{P}}(g):=\mathrm{E}_{\mathrm{P}}[\ell(g(\boldsymbol{x}))], & R_{\mathrm{N}}(g):=\mathrm{E}_{\mathrm{N}}[\ell(-g(\boldsymbol{x}))] \\
R_{\mathrm{U}, \mathrm{P}}(g):=\mathrm{E}_{\mathrm{U}}[\ell(g(\boldsymbol{x}))], & R_{\mathrm{U}, \mathrm{N}}(g):=\mathrm{E}_{\mathrm{U}}[\ell(-g(\boldsymbol{x}))]
\end{array}
$$
where $\mathrm{E}_{\mathrm{P}}, \mathrm{E}_{\mathrm{N}}$, and $\mathrm{E}_{\mathrm{U}}$ denote the expectations over $p_{\mathrm{P}}(\boldsymbol{x}), p_{\mathrm{N}}(\boldsymbol{x})$, and $p(\boldsymbol{x})$, respectively. Since we do not have any samples from $p(\boldsymbol{x}, y)$, the true risk $R(g)=$ $\mathrm{E}_{p(\boldsymbol{x}, y)}[\ell(y g(\boldsymbol{x}))]$, which we want to minimize, should be recovered without using $p(\boldsymbol{x}, y)$ as shown below.

When we have both positive and negative samples in a supervised classification task (i.e., PN classification), the risk can be defined as
\begin{equation}
\begin{aligned}
R_{\mathrm{PN}}(g) & :=\theta_{\mathrm{P}} \mathrm{E}_{\mathrm{P}}[\ell(g(\boldsymbol{x}))]+\theta_{\mathrm{N}} \mathrm{E}_{\mathrm{N}}[\ell(-g(\boldsymbol{x}))] \\
& =\theta_{\mathrm{P}} R_{\mathrm{P}}(g)+\theta_{\mathrm{N}} R_{\mathrm{N}}(g)
\end{aligned}
\label{eq:pn-risk-estimator}
\end{equation}

which is equal to $R(g)$, but $p(\boldsymbol{x}, y)$ is not included.

In PU classification, labeled data for the negative class is absent. Unlabeled data originating from marginal density $p(\boldsymbol{x})$ can be utilized instead. The objective here is to train a classifier using only positive and unlabeled data. While the basic approach is to discriminate $\mathrm{P}$ and $\mathrm{U}$ data~\cite{elkan2008learning}, naively classifying $\mathrm{P}$ and $\mathrm{U}$ data causes a bias. To address this problem, \citeauthor{du2014analysis} proposed a risk equivalent to the $\mathrm{PN}$ risk but without including $p_{\mathrm{N}}(\boldsymbol{x})$. The key idea is to utilize unlabeled data to evaluate the risk for negative samples in the PN risk. Replacing the second term in Eq. \ref{eq:pn-risk-estimator} with \footnote{The equation comes from the definition of the marginal density $p(\boldsymbol{x})=\theta_{\mathrm{P}} p_{\mathrm{P}}(\boldsymbol{x})+\theta_{\mathrm{N}} p_{\mathrm{N}}(\boldsymbol{x})$}
$$
\theta_{\mathrm{N}} \mathrm{E}_{\mathrm{N}}[\ell(-g(\boldsymbol{x}))]=\mathrm{E}_{\mathrm{U}}[\ell(-g(\boldsymbol{x}))]-\theta_{\mathrm{P}} \mathrm{E}_{\mathrm{P}}[\ell(-g(\boldsymbol{x}))] \text {, }
$$
the risk in PU classification (the PU risk) is obtained as follows
$$
\begin{aligned}
R_{\mathrm{PU}}(g) & :=\theta_{\mathrm{P}} \mathrm{E}_{\mathrm{P}}[\widetilde{\ell}(g(\boldsymbol{x}))]+\mathrm{E}_{\mathrm{U}}[\ell(-g(\boldsymbol{x}))] \\
& =\theta_{\mathrm{P}} R_{\mathrm{P}}^{\mathrm{C}}(g)+R_{\mathrm{U}, \mathrm{N}}(g),
\end{aligned}
$$
where $R_{\mathrm{P}}^{\mathrm{C}}(g):=\mathrm{E}_{\mathrm{P}}[\widetilde{\ell}(g(\boldsymbol{x}))]$ and $\widetilde{\ell}(m)=\ell(m)-\ell(-m)$ is a composite loss function.

\subsubsection{PULSAR}
We propose PULSAR, a spatio temporal graph convolution model, where we used the idea of Adaptive Graph Convolution \cite{shi2019two} in context of hand skeletons, allowing us to capture intricate temporal patterns and spatial relationships within the hand movements. In addition to that, to overcome the challenges inherent to the specific problem in hand, we frame the task of detecting Parkinson’s Disease from the finger-tapping task as a Positive Unlabeled (PU) Learning problem where we use a modified risk estimator. 

In a standard spatio temporal graph convolution model (e.g., used for action recognition), the feature vector attached to each vertex only contains the coordinates of the joints, i.e., the first order information of the skeleton data. The hand joint coordinates by itself can be utilized for the PD detection model. However, the second order information, which represent the feature of bones between two joints, or third order information like velocity and acceleration, hasn't been exploited. By leveraging bone vectors derived from joint positions, we capture the relative orientations between joints, which is crucial in assessing motor impairment. In the finger-tapping task, the velocity and acceleration of the tapping motion can provide insights into the patient's motor control abilities. In a healthy individual, the tapping motion is expected to exhibit relatively smooth and controlled movements, resulting in consistent velocity and acceleration profiles. However, in Parkinson's patients, motor impairment can lead to irregular and jerky movements, resulting in variations in velocity and acceleration patterns. 

\begin{figure*}[htb]
  \centering
  \includegraphics[width=0.8\linewidth]{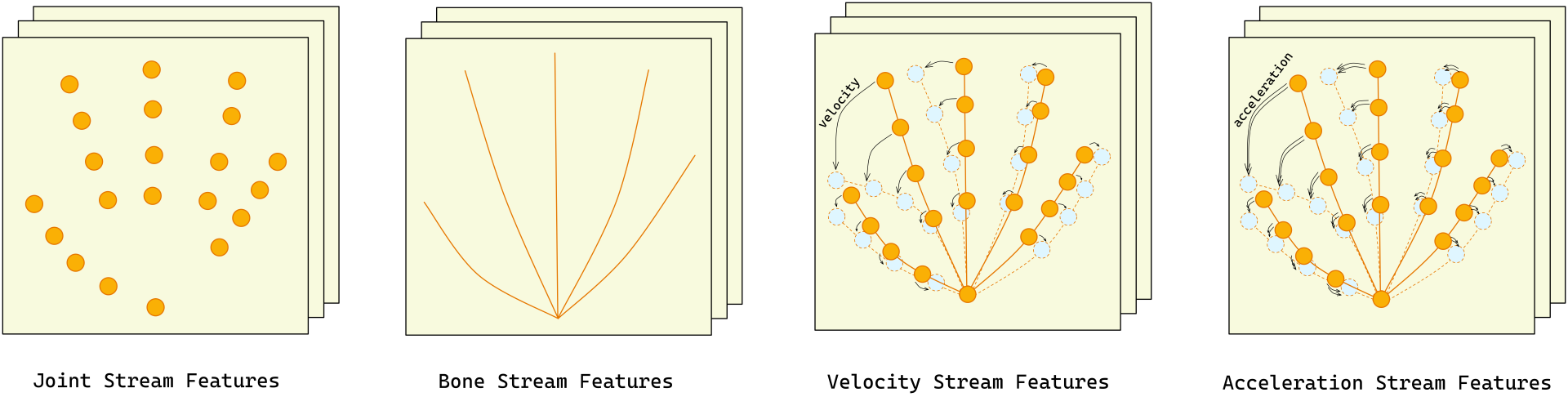}
  \caption{Feature representation of multi stream network of PULSAR}
  \label{fig:multi-stream}
  \vspace{-3mm}
\end{figure*}

To utilize additional information, we integrate streamed attributes involving secondary and tertiary data aspects such as bone-related details, velocity, and acceleration (Figure~\ref{fig:multi-stream}). The bone constructs are established as vectors denoting the directional link between two joints. This directional aspect is preserved by subtracting joint positions. Velocity and acceleration representations are computed by taking the first and second order derivative of joint positions across frames respectively. To avoid mixing the information about the different features, all four models are trained separately.

The model takes in an embedded sequence of joints along with its corresponding adjacency matrix as the input. While the spatial convolution component remains unchanged, the temporal graph convolutional part is executed following the approach proposed by \citeauthor{yan2018spatial}. This involves applying a $K_t \times 1$ convolution on the $C \times T \times V$ feature maps, where $C$ is the number of input features, $T$ is the input time steps, $V$ is the number of vertices in the graph and  $K_t$ is the kernel size of temporal dimension. Both the spatial graph convolutional network (GCN) and the temporal GCN are succeeded by a batch normalization (BN) layer and a Rectified Linear Unit (ReLU) layer. Within each block, a dropout layer with a dropout rate of 0.5 is positioned between the spatial GCN and the temporal GCN. The network we propose is fed by a four-dimensional matrix of shape $N \times C \times T \times V$ where $N$ represents the batch size.

We used two AGCN blocks in our network with increasing output feature dimensions. As a preprocessing step, we included a batch normalization layer to standardize the input data. The output of the model is mapped to the corresponding output classes through a global average pooling layer (GAP) followed by a fully connected layer (FC). The AGCN-block is a fundamental building block of the PULSAR model, which involves a spatial graph convolution followed by a temporal graph convolution with specific kernel sizes. Each of the AGCN blocks has a residual connection for stabilizing the training.

\vspace{-2.5mm}
\subsection{Training Details}
During the training phase, a batch size of 64 and 2 input features (x-y coordinates) are set. The choice of batch size was done through an initial experiment with mini-batch sizes of 32, 64, and 128, where the size of 64 yielded the highest validation accuracy. This decision was also guided by the benefits of smaller batch sizes, which are known to yield smoother minimizers and improved generalization capabilities, as highlighted by \cite{keskar2016large}. The temporal dimension $T$ was chosen to be 80 based on empirical observations, and $V$ was set to 21, which is the number of available hand joints. \texttt{Adam} optimizer was used with an initial learning rate of $1e^{-4}$. A plateau learning rate scheduler was implemented, adjusting according to the validation accuracy, with a reduction factor of 0.5 and a patience of 5 epochs. The specific learning rate and scheduler parameters were fine-tuned across a sequence of experiments, varying the learning rates from $1e^{-3}$ to $1e^{-5}$ and the reduction factors from 0.1 to 0.5. The models were trained for a maximum of 30 epochs.

\begin{table}[!htp]
\begin{center}
\scalebox{0.75}{
\begin{tabular}{|l|p{2.50in}|}\hline
    \textbf{Name}&\textbf{Description}\\\hline
    JS & Model based on joint Stream features\\ \hline
    JS\_PU & Model based on joint Stream features with PU Learning\\ \hline
    JS\_AC & Model based on joint Stream features with Adaptive Convolution\\ \hline
    JS\_AC\_PU & Model based on joint Stream features with Adaptive Convolution and PU Learning\\ \hline
    PULSAR & Model based on joint, bone, velocity and acceleration streams (i.e., multi stream) features with Adaptive Convolution and PU Learning\\ \hline
\end{tabular}
}
\end{center}
\caption{Acronym and short description of baseline models along with PULSAR.}
\label{tab:model_names}
\end{table}
\vspace{-8.5mm}

\section{Results}
In this section we evaluate the performance of PULSAR and compare it with several baseline models as mentioned in Table~\ref{tab:model_names}. All experiments were conducted on PyTorch deep learning framework, 1 Nvidia GTX 1070 GPU and 1 Nvidia Tesla P100 GPU.

\begin{table}[!b]
\begin{center}
\scalebox{0.82}
{
\begin{tabular}{|l|l|l|l|l|l|l|}
\hline
    $Model$ & $Acc$ & $Prec$ & $Rec$ & $F_1^m$ & $F_1^w$ & $AUC$ \\ \hline
    JS     & 60.88 & 61.10 & 61.09 & 60.87 & 60.86 & 61.08 \\ \hline
    JS\_PU & 69.05 & 66.91 & 65.70 & 66.04 & 68.48 & 65.69 \\ \hline
    JS\_AC & 65.90 & 66.05 & 66.06 & 65.90 & 65.91 & 66.06 \\ \hline
    JS\_AC\_PU & 70.24 & 68.32 & 68.02 & 68.16 & 70.12 & 68.02 \\ \hline
    PULSAR   & \textbf{80.95} & \textbf{84.32} & \textbf{82.76} & \textbf{80.87} & \textbf{80.73} & \textbf{82.75} \\ \hline
\end{tabular}
}
\end{center}
\caption{Performance comparison of PULSAR with baseline models on the validation set. Best result for each metric is shown in bold-face. $Acc$: Accuracy, $Prec$: Precision, $Rec$: Recall, $F_1^m$: Macro F1, $F_1^w$: Weighted F1, $AUC$: Area under Receiver Operating Characteristic Curve.}
\label{table:vt}
\end{table}

\vspace{-2.5mm}
\subsection{Performance on Validation Set}
The performance comparison of PULSAR with baseline models on the validation set is shown in Table~\ref{table:vt}. PULSAR has the highest accuracy, precision, recall, F1 (macro as well as weighted) and area under receiver operating characteristic Curve (AUROC). Compared to the joint stream based model, PULSAR enjoys improvements of 33\% and 35\%  in accuracy and AUROC respectively. On the other hand, the inclusion of bone, velocity and acceleration streams alongside joint stream (i.e., from JS\_AC\_PU model to PULSAR) improves the accuracy by 15\%. The benefit of PU learning is also quite clear from the validation results. From JS to JS\_PU model, accuracy increased by 13\%. On the other hand, inclusion of adaptive convolution to the vanilla JS model increased accuracy by 8\%. Overall, all these added components -- multiple streams, PU learning and adaptive convolution -- significantly improved PULSAR over the baseline model with joint stream alone. 

\begin{table}[!b]
\begin{center}
\scalebox{0.82}
{
\begin{tabular}{|l|l|l|l|l|l|l|}
\hline
    $Model$ & $Acc$ & $Prec$ & $Rec$ & $F_1^m$ & $F_1^w$ & $AUC$ \\ \hline
    JS     & 63.46 & 64.20 & 63.56 & 62.92 & 63.05 & 63.56 \\ \hline
    JS\_PU & 64.12 & 66.47 & 64.43 & 62.93 & 62.97 & 64.43 \\ \hline
    JS\_AC & 66.68 & 66.57 & 66.64 & 66.54 & 66.71 & 66.64 \\ \hline
    JS\_AC\_PU & 68.41 & 69.70 & 68.61 & 67.89 & 67.97 & 68.61 \\ \hline
    PULSAR   & \textbf{71.29} & \textbf{71.21} & \textbf{71.22} & \textbf{71.06} & \textbf{71.29} & \textbf{71.22} \\ \hline
\end{tabular}
}
\end{center}
\caption{Performance comparison of PULSAR with baseline models on the independent test set. Best result for each metric is shown in bold-face. All the values are averaged over 20 replicates.}
\label{table:it}
\end{table}

\vspace{-2.5mm}
\subsection{Performance on Independent Test Set}

From the pool of 182 patients left aside for independent testing, we sampled, with replacement, 120 participants' data 20 times. We measured the performance of PULSAR and the baseline models in each of these replicates to determine the average and variation of the various metrics. The average accuracy, precision, recall, F1 (macro as well as weighted) and AUROC are reported in Table~\ref{table:it}. While the performance of PULSAR degraded from validation to the independent test set, it nevertheless outperformed all the other models in the independent testing. And the overall comparative trend among the models also held -- performance improved when either PU learning or adaptive convolution was added to the joint stream based model. When both PU and adaptive convolutions were added, performance was even better. Finally, when bone, velocity and acceleration streams were added to create the PULSAR model, the performance exceeded all the other models. Additionally, the standard deviations of all the performance metrics were less than 3\%, which points to the stability of PULSAR.

\begin{figure}[htb]
  \centering
  \includegraphics[width=\linewidth]{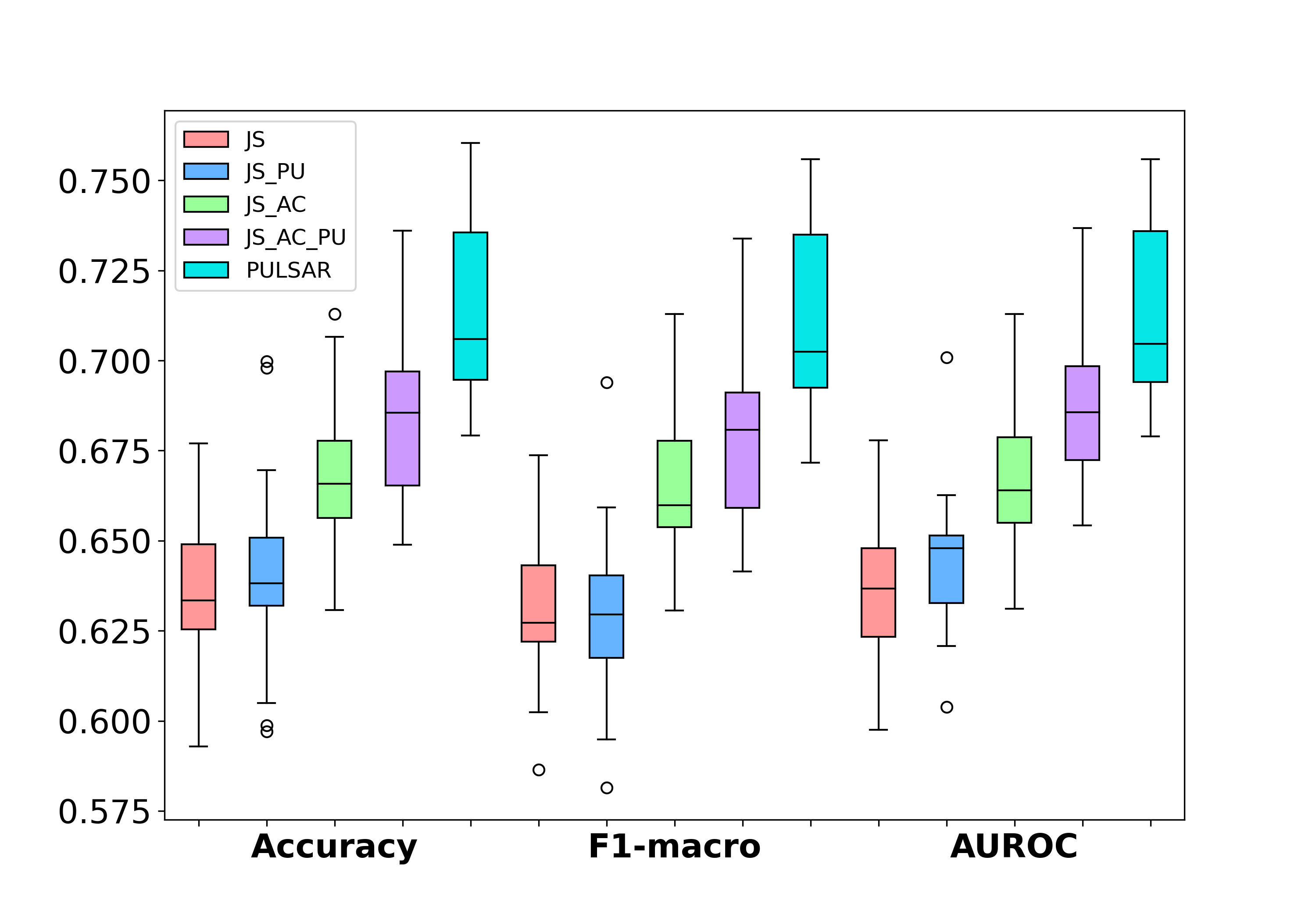}
  \caption{Box and whisker plot for accuracy, macro F1 and AUROC of PULSAR and baseline models.}
  \label{fig:box_plot}
\end{figure}

Figure~\ref{fig:box_plot} shows the box and whisker plot for accuracy, macro F1 and AUROC of the different models. The addition of PU learning to the JS model made the results less dispersed. It thus added stability to the model performance. The addition of adaptive convolution, on the other hand, made the distribution of accuracy normal like. The median line for PULSAR is above the box for each metric of the baseline models. This is an indication that likely PULSAR is statistically significantly different than the other models. This is more rigorously tested in the next section.


\begin{figure}[htb]
  \centering
  \includegraphics[width=0.9\linewidth]{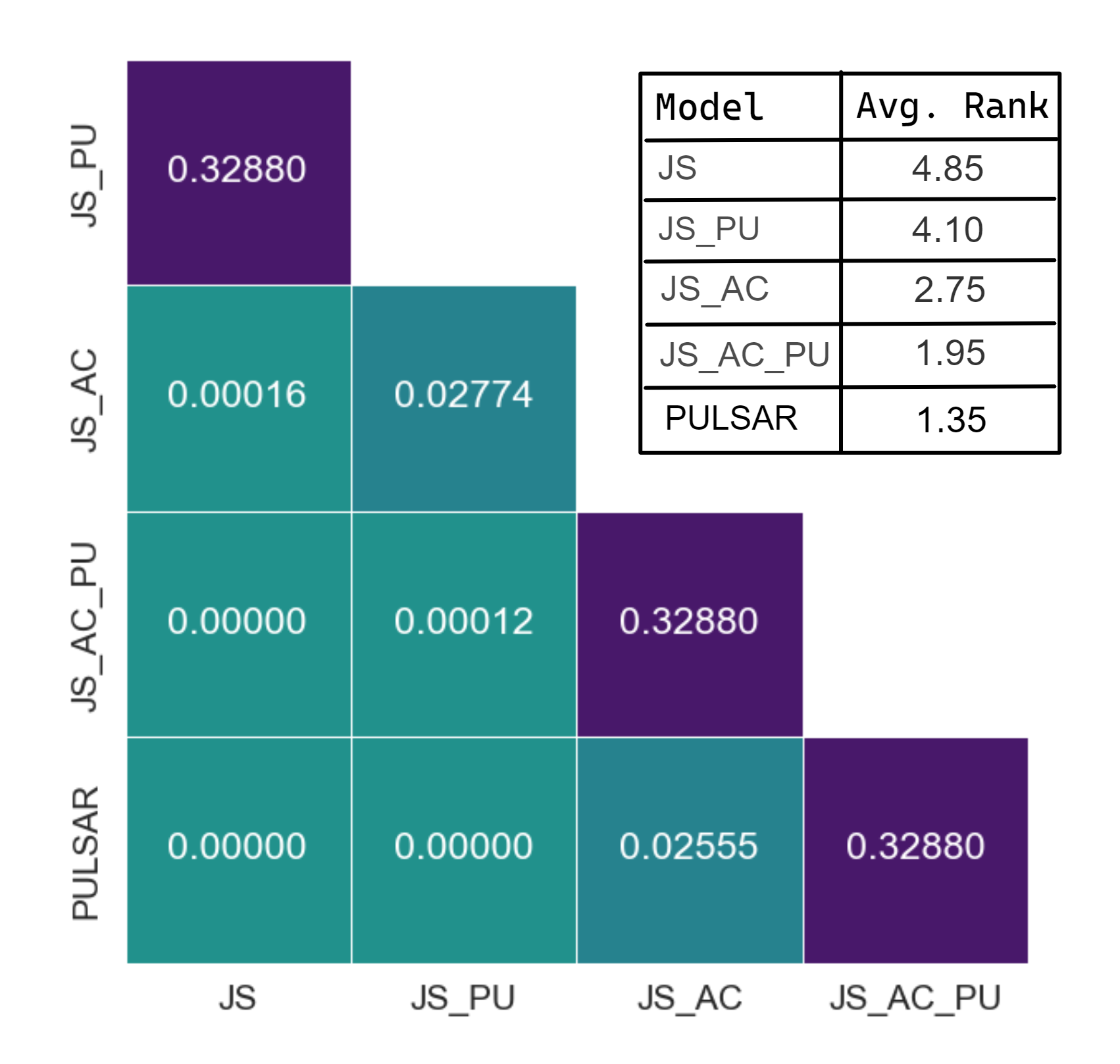}
  \caption{Heatmap of the adjusted $p$-values from the post hoc Holm test on the accuracy metric. Average rank of models based on the Friedman test is shown in the table.}
  \label{fig:pval_heatmap}
\end{figure}

\begin{figure*}[ht]
  \centering
  \includegraphics[width=\linewidth]{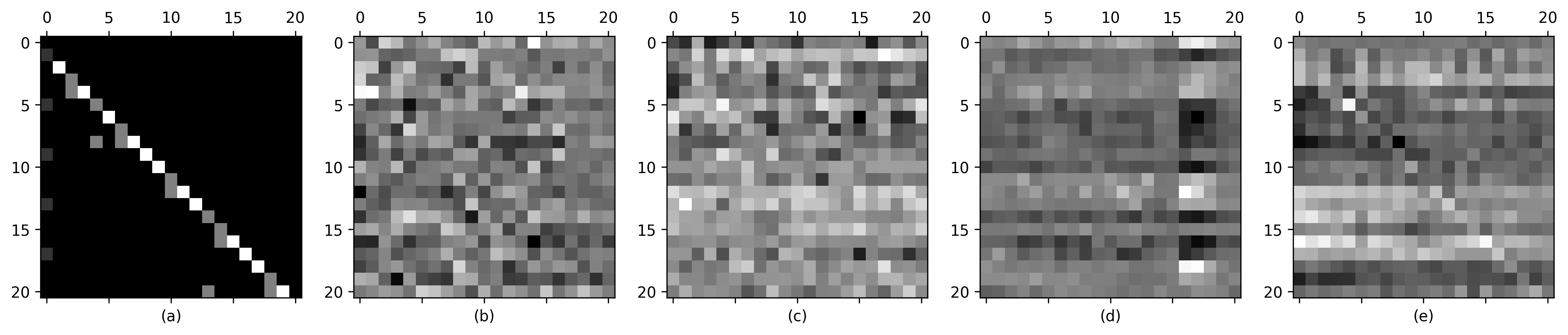}
  \caption{The matrix in (a) is the original adjacency matrix for the second subset. (b) - (e) shows the learned adjacency matrix of joint stream, bone stream, velocity stream and acceleration stream models respectively}
  \label{fig:adj_map}
\end{figure*}

\begin{figure*}[ht]
  \centering
  \includegraphics[width=0.8\linewidth]{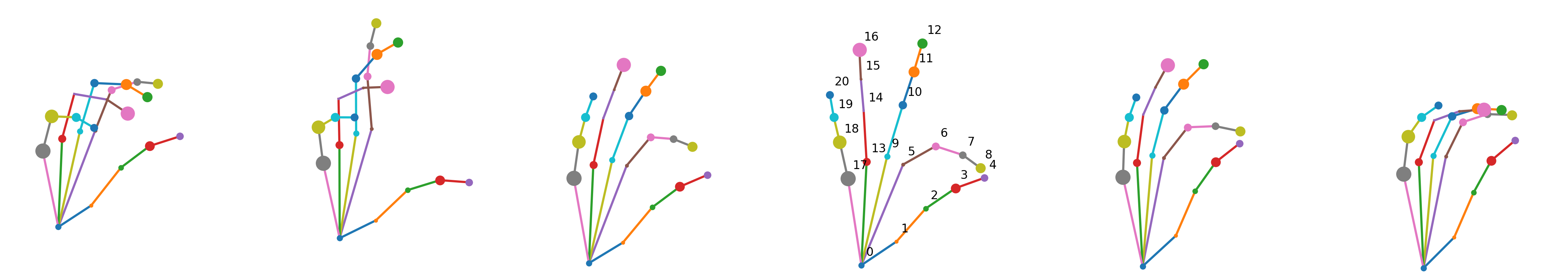}
  \caption{Visualization of the hand key point graph across different frames for one sample. The key points are numbered the same as in Fig. \ref{fig:adj_map}}
  \label{fig:pose}
\end{figure*}

\vspace{-2.5mm}
\subsection{Statistical Significance Test}

We have conducted Friedman non-parametric statistical test (FMT)~\cite{demvsar2006statistical} on the independent test accuracy of the predictors, with the significance level, $\alpha = 0.05$. The Friedman statistic distributed according to Chi-square with 4 degrees of freedom ($df$) was 68.16 ($p$-value 3.58e-11). As this is larger than the critical value (9.488 for $df=4$), the null hypothesis ($H_0$: All the models have identical accuracy) is rejected. Table in Figure~\ref{fig:pval_heatmap} summarizes the average rank of the predictors, where PULSAR comes out on top. Subsequently, we conducted Post hoc Holm test for pairwise comparisons for $\alpha = 0.05$. The adjusted $p$-values are shown as a heatmap in Figure~\ref{fig:pval_heatmap}. Based on these adjusted $p$-values and the results in Table~\ref{table:it}, PULSAR's superiority over the JS, JS\_PU and JS\_AC models is statistically significant. PULSAR is not statistically significantly different than the JS\_AC\_PU model, though it has a better average accuracy over the latter.

\section{Visualization of the learned graphs}

 Fig. \ref{fig:adj_map} illustrates the adjacency matrix learned by our models. The gray scale intensity of each element in the matrix represents the strength of the connection between the corresponding joints. The first image (a) is the original adjacency matrix for the second subset employed in PULSAR. The remaining four images (b-e)  correspond to the adaptive adjacency matrices learned independently by each of our four stream models. Clear differences can be observed between the fixed original adjacency matrix and the learned matrices. The learned graph structure exhibits more flexibility and is not constrained to the physical connections of the hand joints. Each of the stream models adjusts the connections differently according to the distinctive patterns it extracts from the respective input features.

 To further visualize the learned connections, we plot the hand key points and the strengths with respect to the wrist for the velocity stream model. Figure \ref{fig:pose} visualizes the hand skeleton graph for the velocity stream model. The skeletons are plotted based on the physical connection between joints. Each circle represents one joint, and its size represents the strength of the connection between that joint and the wrist according to the adjacency matrix in Fig. \ref{fig:adj_map}. From the adjacency matrices in Fig. \ref{fig:adj_map}, we saw that each model learned a different set of connections specific to that streamed feature. In Fig. \ref{fig:pose}, we observe significantly larger circles for key points 16, 17 and 18, which is also apparent from the learned adjacency matrix that shows that the strength of these key points is much higher compared to other key points. We argue that the relative velocity pattern of these key points contains the most distinctive patterns for identifying PD patients. By adaptively learning the graph structure, the model is able to focus on the most relevant joints and motion patterns. This allows the model to extract distinctive features tailored to the Parkinson's disease classification task, rather than relying on predefined skeletal constraints. Through learning, the model determines which relationships in the hand movement data are most useful to leverage for optimal discriminative performance.

\vspace{-2.5mm}
\section{Discussion}
PULSAR employs adaptive graph convolutional neural network for PD screening from the videos of the finger-tapping task. For graph construction, we have used our knowledge of human anatomy and constrained it with geometry. PULSAR's performance demonstrate the potential of leveraging advanced deep learning techniques to facilitate PD screening. \citeauthor{rizzo2016accuracy} showed that the accuracy of clinical diagnosis of PD is 73.8\%, when performed mainly by non-experts in movement disorders (i.e., general neurologists, geriatricians, or general practitioners). In our experiments, PULSAR achieved 80.95\% accuracy in validation set and a mean accuracy of 71.29\% (2.49\% standard deviation) in the independent test set. This shows the potential of PULSAR to be integrated into the health care system, provided that due consideration has been given to data security, privacy and ethics.  

Early diagnosis of PD followed by regular physician visits is crucial to improve the quality of life as well as life expectancy of a PD patient~\cite{fujita2021effects}. As the projected total economic burden in US due to PD surpasses \$79 billion by 2037~\cite{yang2020current}, early PD diagnosis is not only crucial for the patients themselves, but also for the sustainability of the overall healthcare system. However, early PD symptoms can be mistaken for those of normal aging. Therefore, an undiagnosed PD patient may not make the first visit to a neurologist for a long time. This is specially so in healthcare settings with an extreme scarcity of neurologists. PULSAR can have a significant impact in public health by making the screening easily accessible to anyone anywhere with webcam and computer to record the fingar-tapping task, and internet connection to send to video over. The AI model can then automatically assess the symptoms and refer the patient to a neurologist if necessary.

To train PULSAR, we augmented the dataset by flipping the videos. It is also worthwhile to introduce various types of noise, such as, blurriness, and heterogeneous lighting. These can occur naturally in home setting, and the model needs to be robust against such variations in videos. Also, PD symptoms can be suppressed for patients on medication, which can confuse the model. When medication related information is available, the training should incorporate that to improve the model's learning. This can be an interesting future direction. In future, we plan to improve PULSAR's performance using a larger training set. In addition, we envision extending our proposed technique to assess other movement disorders like ataxia and Huntington's disease. Our work shows promise in advancing PD screening accessibility and warrants further research to improve its robustness and scope. However, note that PULSAR is a support tool, not a substitute for clinical evaluation.



\bibliography{aaai22}

\end{document}